\documentclass{llncs}

\let\llncssubparagraph\subparagraph
\let\subparagraph\paragraph
\usepackage[compact]{titlesec}
\let\subparagraph\llncssubparagraph

\usepackage{balance}
\usepackage{float}
\usepackage{algorithm}
\usepackage{algpseudocode}
\usepackage{cite}
\usepackage{amsmath,amssymb,amsfonts}
\usepackage{graphicx}
\usepackage{textcomp}
\usepackage{xcolor}
\def\BibTeX{{\rm B\kern-.05em{\sc i\kern-.025em b}\kern-.08em
    T\kern-.1667em\lower.7ex\hbox{E}\kern-.125emX}}

\usepackage{kantlipsum,setspace,titlesec}

\usepackage{tikz}
\usepackage{amsmath}
\usepackage{amssymb}
\usepackage{xspace}
\usepackage{xurl}
\usepackage{filecontents}
\usepackage{cleveref}
\usepackage{comment}
\usepackage{graphicx}
\usepackage{float}

\usepackage[shortlabels]{enumitem}
\setlist[enumerate]{nosep}
\usepackage{mathtools}
\usepackage[hang,flushmargin]{footmisc} 
\usepackage{xcolor}
\usepackage{kantlipsum,setspace,titlesec}
\usepackage{subcaption}

\usepackage{xpatch}

\newcommand{\name}[0]{\textsc{ScaleViz}\xspace}
\newcommand{\visrec}[0]{\texttt{Vis-Rec}\xspace}

\newcommand{\Random}[0]{\texttt{Random}\xspace}

\newcommand{\Greedy}[0]{\texttt{Greedy}\xspace}

\newcommand{\Sample}[0]{\texttt{Sample}\xspace}
\usepackage{graphbox}
\usepackage{nicefrac}

\usepackage{amsmath}

\usepackage{amsmath}
\usepackage{mathrsfs}
\usepackage{comment}

\usepackage{bm}
\usepackage{bbm}
\usepackage{amssymb}
\usepackage{enumitem}
\usepackage{todonotes}
\usepackage{authblk}
\usepackage[shortlabels]{enumitem}
\setlist[enumerate]{nosep}
\setlist{nolistsep,leftmargin=2.0mm}

\begin{document}

\title{\name: Scaling Visualization Recommendation Models on Large Data}


\author{Ghazi Shazan Ahmad\inst{1}\textsuperscript{\dag} \and
Shubham Agarwal\inst{2} \and
Subrata Mitra\inst{2}\thanks{Corresponding author (subrata.mitra@adobe.com)} \and 
Ryan Rossi\inst{2} \and
Manav Doshi\inst{3}\textsuperscript{\dag} \and Vibhor Porwal\inst{2} \and Syam Manoj Kumar Paila\inst{4}\textsuperscript{\dag}}


\institute{\textsuperscript{1}IIIT B,  \textsuperscript{2}Adobe Research, \textsuperscript{3}IIT B, \textsuperscript{4}IIT KGP}

\renewcommand\thefootnote{\dag}
\footnotetext[1]{Work done during internship at Adobe Research\label{adobe}}

\authorrunning{Ahmad, G.S. et al.}
%

\maketitle

\pagenumbering{arabic}
\thispagestyle{plain} 
\pagestyle{plain}

\begin{abstract}
Automated visualization recommendation (\visrec) models help users to derive crucial insights from new datasets. Typically, such automated \visrec models first calculate a large number of statistics from the datasets and then use machine-learning models to score or classify multiple visualizations choices to recommend the most effective ones, as per the statistics. 
However, state-of-the-art models rely on a very large number of expensive statistics and therefore using such models on large datasets becomes infeasible due to prohibitively large computational time, limiting the effectiveness of such techniques to most large real-world datasets. 
In this paper, we propose a novel reinforcement-learning (RL) based framework that takes a given \visrec model and a time budget from the user and identifies the best set of input statistics, specifically for a target dataset, that would be most effective while generating accurate enough visual insights.
We show the effectiveness of our technique as it enables two state of the art \visrec models to achieve up to $10$X speedup in \textit{time-to-visualize} on four large real-world datasets. 

\end{abstract}


\section{Introduction}

As more and more data is being collected from various sources, users often encounter data that they are not familiar with. A dataset can contain numerous columns, both numerical and categorical, with multiple categories.  It is a daunting task to even decide how to dissect and plot such data to reveal any \textit{interesting} insights.
Visualization recommendation (\visrec) techniques~\cite{hu2019vizml,10.1145/3447548.3467224} help to automatically generate, score, and recommend the most relevant visualizations for a dataset and can improve productivity by reducing the time required by analysts to first find interesting insights and then visualize them.

Automated \visrec techniques typically work through the following steps.
First, they calculate various statistics, as much as up to $1006$ number of statistics as used by Qian et al. in ~\cite{10.1145/3447548.3467224} per column from the data to capture overall statistical landscape of the data. Second, these statistics are used as features to score prospective visualization configurations (i.e. combination of columns, aggregates, plot types etc.) in a supervised learning setup. Finally, queries are issued against the actual data to populate the top recommended visualization charts.
A significant number of prior works \cite{7457691, 10.1145/3588710, hu2019vizml, harris2021insight, vartak2017towards} focused on perfecting the visualization recommendation technique, which evolved from initial algorithmic approaches to most recent deep-learning based approaches~\cite{luo2018deepeye, 10.1145/3447548.3467224, hu2019vizml}. 
Further, Qian et al. \cite{10.1145/3447548.3467224} extended these techniques to address the problem of how to generalize these models on unseen datasets having completely different schema structure and data distributions. 
The way such generalization work is that the neural network learns the importance of different visualizations at much abstract level by extracting a large number of higher order statistical features extracted from the data. 
However, prior works did not address another very important problem, which is the scalability of these algorithms on datasets with large number of columns and/or rows.
Real world datasets can be huge, having several hundreds of millions or even several hundreds of billions of rows.
Calculating large number of statistical features on such large datasets is intractable by the state-of-the-art (SOTA) visualization recommendation algorithms. 
For example, in Qian et al. \cite{10.1145/3447548.3467224} collects $1006$ various higher-order statistics \textit{per column} of the dataset, which itself can have a large number of columns. On top of that, they calculate multi-column statistics to capture dependency, correlation and other such properties. In Fig. \ref{fig:cdf} we show the CDF of computation time for different statistical features that are needed by SOTA \visrec models MLVR \cite{10.1145/3447548.3467224} and VizML \cite{hu2019vizml} for 4 datasets of different sizes. Table \ref{tab:features_num} lists the number of rows and columns for each dataset and total number of statistical features that needs to be computed for MLVR and VizML for each dataset. 
%
%
Calculating so many statistics on large datasets makes the very first step of the typical visualization recommendation pipeline infeasible and unscalable. 

Now, there can be two ways to overcome this problem:
\begin{itemize}
    \item First option could be to drop certain statistics to reduce the computation. \textit{But which ones?} These statistics are basically the features to the core visualization recommendation model. Which statistics are important and carry important signals that would make a particular combination of columns and visualization style interesting - is very dataset dependent. A statistics that is very computationally intensive might carry significant importance for one dataset and might not be relevant for another dataset. Indiscriminate dropping of certain statistics or identifying the important statistics based on few datasets and extending that decision to other datasets, can lead to poor quality output.
    
    \item Second option could be to take a small sample of the data, on which calculation of large number statistics is tractable, and then generate the visualization recommendations based on that sample. However, for massive amounts of data, such sample has to be a tiny fraction for the existing \visrec pipeline to work and such a tiny sample may not be representative of the complete data. Therefore, the visualization recommendations generated on the sample can be completely misleading or inaccurate. 
\end{itemize}

To overcome these drawbacks of naive ways to speed-up visualization recommendation generation, in this paper we present a framework, called \name (code\footnote{\url{https://anonymous.4open.science/r/ScaleViz-30DB}}), that takes such a generalized \visrec model and \textit{customizes} it for a given dataset so that we can produce visualization recommendations at large scale, for that dataset.
\name does this by through the following steps:
(1) It profiles the computational cost of calculating statistics for each statistics that are needed by the generalized model on a few samples of data of different sizes.
(2) It uses regression models to extrapolate that cost to the size of the full dataset.
(3) It uses a \textit{budget-aware} Reinforcement-Learning (RL) based technique to identify the most crucial features from the original \visrec model --- using multiple samples containing a very small fraction from the original dataset.
(4) Finally, \name only calculates these selected statistical features from the full dataset and produces the visualization recommendations using the given model.
In summary, we make the following contributions:
\begin{enumerate}
    \item We propose a framework that enables a \visrec model to generate accurate enough insights for a target large-scale dataset within a chosen time budget.
    \item We formulate the problem as a budget-aware Reinforcement Learning problem that incrementally learns the most useful statistical features from a large-scale dataset for the target model.  
    \item Our evaluations with 2 recent ML-based \visrec models~\cite{hu2019vizml, 10.1145/3447548.3467224} and with 4 large public datasets show that \name can provide upto $10$X speedup in producing accurate enough visualization recommendations.

\end{enumerate}

\begin{table}[!t]
\vspace{0pt}
    \begin{minipage}{.62\linewidth}
    \vspace{0pt}
        \centering
        \begin{subfigure}[t]{0.49\textwidth}
            \centering
            \includegraphics[width=\textwidth]{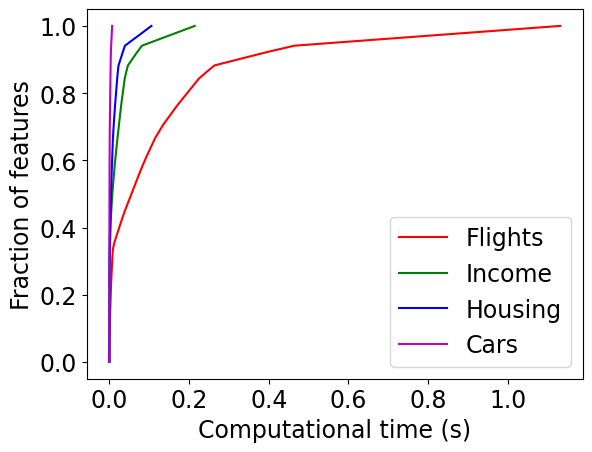}
            \caption{VizML~\cite{hu2019vizml}}
            \label{fig:y equals x}
        \end{subfigure}
        \begin{subfigure}[t]{0.49\textwidth}
            \centering
            \includegraphics[width=\textwidth]{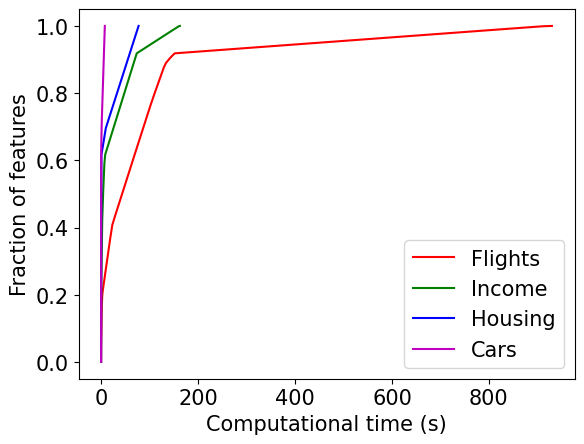}
            \caption{MLVR~\cite{10.1145/3447548.3467224}}
            \label{fig:growth_fig}
        \end{subfigure}
        \captionof{figure}{CDF of computation time (in seconds) of various features on datasets of different sizes. With increase in dataset size, the computation time of complex features drastically increases.}
        \label{fig:cdf}
    \end{minipage}%
    \vspace{0pt}
    \begin{minipage}{.38\linewidth}
    \vspace{0pt}
        \centering
        \scalebox{0.84}{
        \begin{tabular}{|l|l|l|}
            \hline
            \textbf{Datasets}  & \textbf{VizML} & \textbf{MLVR} \\ 
            \hline
            Flights  ($1M, 12$) & 972 & 12072 \\ 
            Income  ($200k, 41$) & 3321 & 41246 \\ 
            Housing  ($20k, 10$) & 810 & 10060 \\ 
            Cars  ($10k, 9$) & 729 & 9054 \\ 
            \hline
        \end{tabular}
        }
        \caption{Number of statistical features that are needed by MLVR and VizML for 4 different datasets with (\# rows, \# columns).}
        \label{tab:features_num}
    \end{minipage}
    \label{table:features_vs_time_fig_1}
\end{table}



\section{Related Works}

Several prior works \cite{10.1145/3447548.3467224, 7457691, 10.1145/3588710, hu2019vizml, harris2021insight, vartak2017towards, ding2019quickinsights, idreos2015overview} targeted visualization recommendations to help insight discovery.
But scalability of such technique when handling large datasets were not addressed and is the focus of this paper.
As recent \visrec models use large number of statistical features from the data, feature selection literature is also related to our work.
Prior works by Li et al. \cite{li2017feature}, Deng et al. \cite{deng2012feature}, and Farahat et al. \cite{farahat2011efficient}, have employed decision trees or greedy-based approaches. Some researchers have explored reinforcement learning techniques for intelligent feature selection, as seen in the works of Kachuee et al. \cite{kachuee2019opportunistic} and others \cite{DBLP:journals/corr/abs-2101-09460}. However, these approaches are not applicable to our setting as for \visrec models, the crucial features are often dependent on the particular statistical characteristics of the target data and so can not be selected at the training time.

\section{Problem Formulation}

In this section, we first formally define the problem of budget-aware visualization recommendation generation. 

Let $\mathcal{P}$ be a target \visrec model that a user wants to apply on a large tabular dataset $\mathcal{D}$.
Let $\mathcal{D}$ consists $m$ columns and $r$ rows. 
Let $\mathcal{F}$ be the feature space for dataset $\mathcal{D}$ based on statistical features used in the model $\mathcal{P}$.
As \visrec models calculate a large number of different statistics from each column, let us denote the number of statistical features computed from each column be $n$. 
Let $f_{ij}$ denote the $j$-th feature for $i$-th column, where $i \in \{1, \ldots , m\}$ and $j \in \{1, \ldots ,n\}$.

We introduce the cost function $c^{k} : \mathcal{F} \to \mathbb{R}^{m\times n}$, quantifying the \textit{computational time} required to calculate each of the features based on a $k$ fraction from $\mathcal{D}$ (i.e. such fraction will consist of $1/k$ rows of $\mathcal{D}$). 
Notably, $c^{1}$ serves as the cost function for the entire dataset $\mathcal{D}$, and for brevity, we use $c^1$ denoted as $c$ throughout the paper.

To formalize the problem, we frame the statistical feature selection as an optimization task. Let $\theta : \mathcal{F} \to \{0, 1\}^{m \times n}$ be a function mapping features to binary acquisition decisions. $\theta(f) \odot f$ gives a subset of features by ignoring the masked features, where $f \in \mathcal{F}$ and $\odot$ is the Hadamard operator which calculates the product of two matrices of the same dimension. $\mathcal{L}$ is a loss function which compares the output of the model on two different input feature set. The objective is to find the feature mask minimizing the error in the model's prediction while ensuring the total cost of selected features adheres to the budget $\mathcal{B}$:
\begin{equation}
    \min_\theta \mathcal{L}[\mathcal{P}(\theta(f) \odot f) - \mathcal{P}(f)], \text{ subject to: } \sum\limits_{i, j} \theta(f) \odot c(f) \leq \mathcal{B}
\end{equation}


Here, the budget $\mathcal{B}$ is constrained by the total computational cost of features calculated on the complete dataset:
\begin{equation}
    \mathcal{B} \leq \sum\limits_{i, j} c(f)
\end{equation}

Note, in this formulation, we use $\mathcal{B}$, that is \textit{time-to-compute} visualization recommendations as the constraint, because it is intuitive for users to specify a time-budget. 
Alternatively, we could also make this constraint relative to the size of $\mathcal{D}$. In that case, $\mathcal{B} \leq r \times \sum\limits_{i, j} c(f)$ where $r$ is a particular user-specified fraction of the statistical feature computation time for the base \visrec model.  

\begin{figure}[H]
    \centering
    \includegraphics[width=\textwidth]{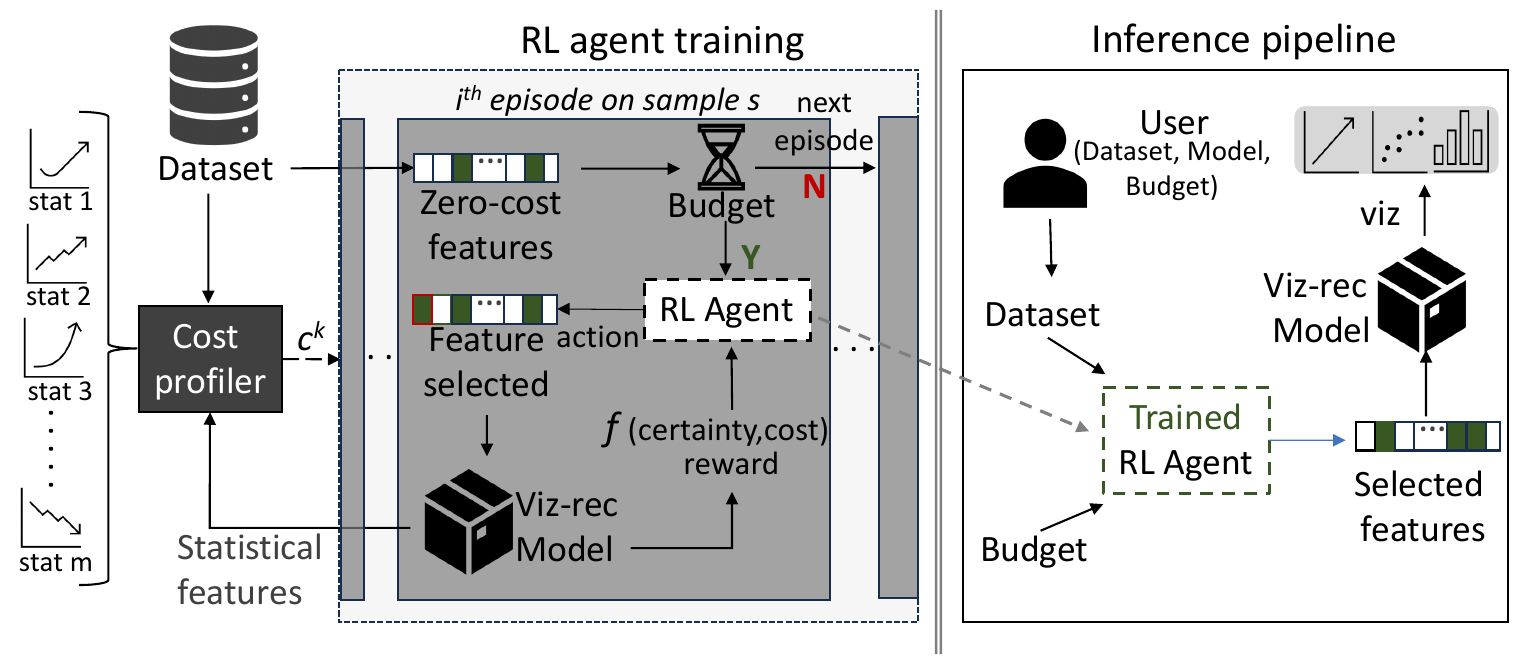}
    \caption{Pipeline overview: The cost profiler estimates the computational cost for computing statistics. RL agent training begins with a set of zero-cost features. Within each episode, the agent dynamically acquires features until the budget is exhausted. The $Q$-value is estimated using rewards which is based on increased certainty in recommendations with newly acquired features, considering acquisition costs. This iterative process continues until error converges below a certain error value. Once trained, in the inference pipeline, the RL agent now selects features for the specified budget, tailored to the dataset and model.} 
    \label{fig:pipeline}
\end{figure}

\section{Proposed Solution}

We approach the above problem as a scenario where decisions are made sequentially over time and model the problem as a reinforcement learning problem. The overall pipeline, as shown in Fig. \ref{fig:pipeline}, consists of a \textbf{Cost profiler}, which employs polynomial regression to estimate the computational cost of computing statistics across varying dataset sizes. This estimation is crucial for predicting costs without actually computing them. Subsequently, the \textbf{RL agent training} module teaches the agent to acquire features under budget constraints across increasing data samples. Once trained, the \textbf{Inference pipeline} utilizes the RL agent to select features for the given budget, computing only the learned subset of features on the entire dataset to obtain model predictions. We provide a detailed description of the two main components and also describe the RL agent training algorithm.

\subsection{Cost Profiling}

The Cost Profiler module profiles the computation time (cost) of each statistical feature across varying dataset sizes. It collects data points to estimate the computation cost for each feature on larger datasets without actual computation.

Given the dataset $\mathcal{D}$, the cost function $c^{k}$ is obtained for $|k|$ fractions of the dataset, denoted as $\{c^{k_1}, c^{k_2}, \ldots , c^{|k|}\}$. For each feature $f_{ij}$, the goal is to predict its cost $c_{ij}$ on the full dataset. Some features, such as \textit{column types, number of categories in a column, max-min value in a column}, exhibit zero-cost, implying their cost remains constant with growing record sizes, i.e $c_{ij} = 0$. For other features, assuming polynomial growth of feature costs with dataset size (as proved in \cite{wang2016statistical}).

\begin{algorithm}[thb]
\caption{RL algorithm in \name to identify important features}
\begin{algorithmic}[1]

    \Statex Given a dataset $\mathcal{D}$, budget $\mathcal{B}$ and a model $\mathcal{P}$

    \Function{\name}{$\mathcal{B}$, $\mathcal{D}$, $\mathcal{P}$}:
    \State $S \gets [S_1, S_2, S_3, ...., S_{
    |S|}]$  $\text{($|S_{k+1}| > |S_{k}|$ $\forall k \in [1, |S| - 1]$)}$
    \For {sample $S_k$ \textit{in the samples set} S}
        \State $x^{k, t} \gets$  $[f_{11}^{k, t}, f_{12}^{k, t}, \ldots , f_{1n}^{k, t}, f_{21}^{k, t} , \ldots f_{2n}^{k, t}, \ldots, 
            \ldots f_{m1}^{k, t}, \ldots , f_{mn}^{k, t} ]$
        \State $\Tilde{y_{k}} \gets$ \text{score predicted by $\mathcal{P}$ on all features}, $terminate \textunderscore flag \gets False$
        \While{not $terminate \textunderscore flag$}
            \If{$random$ in $[0,1) \leq Pr_{\text {rand}}$} 
                \State $ij \gets$ \text{index of a randomly selected unknown feature}
            \Else
                \State $ij$ $\gets$ $Q(x^{k, t})$ $\text{(index of the feature with the maximum Q value)}$
            \EndIf
            \State $x^{k, t+1} \gets \text{acquire $f_{ij}$ and unmask it}$
            \State $\mathcal{P}(x^{k, t}) \gets \text{score predicted using the feature set $x^{k, t}$}$
            \State $total \textunderscore cost \gets total \textunderscore cost +\boldsymbol{c}_{\boldsymbol{ij}}$ 
            \State $r_{ij}^{k, t} \leftarrow \frac{\left\|\operatorname{\mathcal{P}}\left(\boldsymbol{x}^{k,t}\right)-\operatorname{\mathcal{P}}\left(\boldsymbol{x}^{k, t+1}\right)\right\|}{\boldsymbol{c}_{ij}} $
            \State \text{push} $\left(\boldsymbol{x^{k, t}}, ij, \boldsymbol{x^{k, t+1}}, r_{ij}^{k, t}\right)$ \text{into the replay memory}
            \State $t \leftarrow t+1$
            \If{$total \textunderscore cost \geq B$}
                \State $terminate \textunderscore flag \gets True$
                \State $\hat{y_k} \gets \mathcal{P}(x^{k, t})   \text{predicted score on subset of features $x^{k, t}$}$
            \EndIf
            \State \text{loss} $\gets$ $\mathcal{L}(\Tilde{y_k}$,$\hat{y_k})$
            \If{$update \textunderscore condition$}
                \State \text{train \textunderscore batch} $\gets$ \text{random mini-batch from the replay memory}
                \State \text{update (Q, target Q) networks using train batch }
            \EndIf
                
        \EndWhile
        \If{$\epsilon > loss$}
            terminate loop
        \EndIf
    \EndFor
    \EndFunction
\end{algorithmic}
\label{algo:full}
\end{algorithm}

\subsection{RL agent}
We use an RL agent based framework to learn feature acquisition under budget constraints. Each episode consists of the agent choosing the important subset of features for a sample $S_k$. We define the state of the agent for an episode $k$ as the feature set acquired by it so far in an episode (i.e $x^{k, t} = {\theta^{k, t}(f) \odot f}$), where $\theta^{k, t}$ is the mask of the features at time $t$. The action $a_{k, t}$ of the agent is to select a feature which has not been masked in the feature set (i.e $x^{k, t}$). 

At every step $t$, the agent selects a feature $ij$ and masks that feature as selected. The agent moves to the next state, which is $x^{k, t+1} = {\theta^{k, t+1}(f) \odot f}$).
A cost of $c_{ij}$ is deducted from the remaining budget for choosing the feature. 
The reward for an action $a_{k, t}$ is calculated as the absolute change in the score before and after acquiring the feature, $ij$ with a penalty of $c_{}$.

\begin{equation}
    r_t = \frac{\left\|\operatorname{\mathcal{P}}\left(\boldsymbol{x}^{k, t}\right)-\operatorname{\mathcal{P}}\left(\boldsymbol{x}^{k, t+1}\right)\right\|}{\boldsymbol{c}_{ij}}
\end{equation}

We use the technique of double deep $Q$-learning with experience replay buffer \cite{mnih2015human} to train the RL agent. The agent explores the feature space with a $\epsilon$-greedy approach, with the probability of exploration decaying exponentially. The architecture of the $Q$-networks is a feed-forward neural network with three layers of sizes $[512, 128, 64]$.

Algorithm \ref{algo:full} describes the training procedure for the RL agent, designed for cost-effective feature acquisition. 
The process initiates with the agent receiving a dataset $\mathcal{D}$, a pre-defined budget $\mathcal{B}$ and a \visrec model $\mathcal{P}$. The dataset is sequentially explored through a series of samples. The algorithm initializes by setting an initial exploration probability $Pr_{\text{rand}}$ and a termination threshold $\epsilon$. In each episode, the agent learns the important subset of features for a particular sample $S_k$. Every episode starts with the same budget, $\mathcal{B}$ and the size of the samples keeps increasing with the number of episodes. The RL agent starts with a \textit{zero-cost} feature set and keeps acquiring features till it runs out of budget.  At every step of an episode, the agent chooses to explore randomly or exploit the current knowledge by selecting the feature with the maximum $Q$-value. The tuple (state, action, next state, reward) is pushed into the experience replay buffer. The $Q$ and the target-$Q$ networks are periodically updated using the tuples from the buffer. The process is terminated when the loss for an episode falls below the threshold $\epsilon$. The increasing size of the samples across episodes helps the agent to exploit the learned behavior of the model on a larger sample. This is particularly important because, we ultimately want the agent to predict the important features on the full dataset which it has not been trained on.  

The RL agent ultimately selects the important and highly sensitive statistical features for the target base \visrec model $\mathcal{P}$ from a given dataset $\mathcal{D}$.


\section{Evaluations}

\subsection{Experimental Setup}
We use an NVIDIA GeForce RTX 3090 GPU and a 32-core processor for all experiments. PyTorch, scikit-learn, and pandas were used for both training of the agent and running \visrec models. For \visrec input features, non-selected features were imputed based on a smaller 0.01\% sample. We use an exponential decay exploration probability which starts with probability of $1.0$ and eventually reaches $0.1$. A batch size of $128$ is used to randomly sample experiences from the replay buffer. The agent's action space was normalized to facilitate efficient Q-network training. The training of both the Q and target Q networks employed the Adam optimization algorithm and Mean Squared Error (MSE) loss for effective convergence.
\vspace{-1em}
\subsubsection*{\visrec Models}
\begin{enumerate}
    \item \textbf{VizML\cite{hu2019vizml}:} 
    VizML provides visualization-level and encoding-level prediction tasks using 81 column-level features, which are aggregated using 16 functions for predicting visualizations. In our experiments, the RL agent selects column-level features during training, which are then aggregated and fed into the VizML model. We use cross-entropy loss to calculate the error introduced due to selection of subset of features.


    \item \textbf{MLVR\cite{10.1145/3447548.3467224}:} MLVR recommends the top-$k$ visualizations for a given dataset and set of visualizations. It predicts visualization probabilities by leveraging $1006$ column-level features. This approach becomes computationally challenging with a high number of columns. In our experiments, we use the mean-squared loss of the prediction scores on the top-$k$ visualization configurations given by the model using the full feature set to calculate the error. 
\end{enumerate}


\vspace{-1em}

\subsubsection*{Datasets}
\begin{enumerate}
    \item \textbf{Flights:}\footnote{\url{https://www.kaggle.com/datasets/mexwell/carrier-dataset}} On-time performance of domestic flights operated by large air carriers in the USA. It comprises approximately $1$ million rows and $12$ columns.
    \item \textbf{Income:}\footnote{\url{https://www.kaggle.com/datasets/manishkc06/usa-census-income-data}} USA Census Income data, with $200k$ rows and $41$ columns.
    \item \textbf{Cars:}\footnote{\url{ https://www.kaggle.com/datasets/mrdheer/cars-dataset}} Features of vehicles, including mileage, transmission, price, etc. The dataset consists of $10k$ rows and $9$ columns.
    \item \textbf{Housing:}\footnote{\url{https://www.kaggle.com/datasets/ashydv/housing-dataset}} Home prices based on factors like area, bedrooms, furnishings, proximity to the main road, etc. It contains around $20k$ rows and $10$ columns.
\end{enumerate}

\vspace{-1em}

\subsubsection*{Baselines}
We compare \name with the following baselines.
\begin{enumerate}
    \item \textbf{\Random:} Features are randomly selected through uniform sampling until the budget is exhausted, forming the set of features for the \visrec model.
    \item \textbf{\Greedy:} Features are chosen using a greedy technique inspired by \cite{xu2012greedy} until the budget is exhausted, and then passed to the \visrec model.
    \item \textbf{\Sample:} Features are computed on $1\%, 2\%, 3\%,$ and $5\%$ uniform samples of the dataset. This baseline approach allows calculation of all the statistics using a small sample, which can be passed to the \visrec model.
\end{enumerate}



\begin{figure}[H]
     \centering
     \begin{subfigure}[b]{0.33\textwidth}
         \centering
         \includegraphics[width=\textwidth]{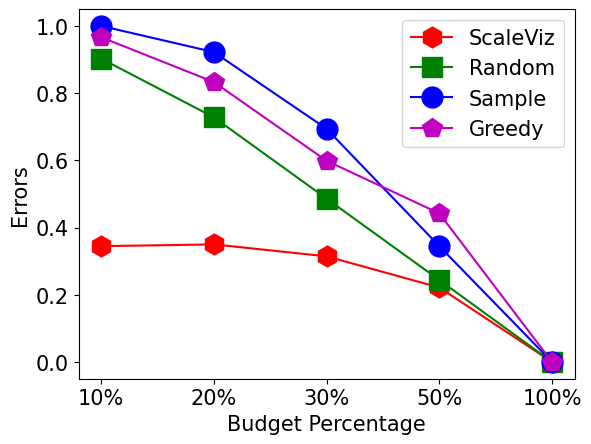}
         \caption{Flights dataset.}
         \label{fig:y equals x}
     \end{subfigure}
     \begin{subfigure}[b]{0.33\textwidth}
         \centering
         \includegraphics[width=\textwidth]{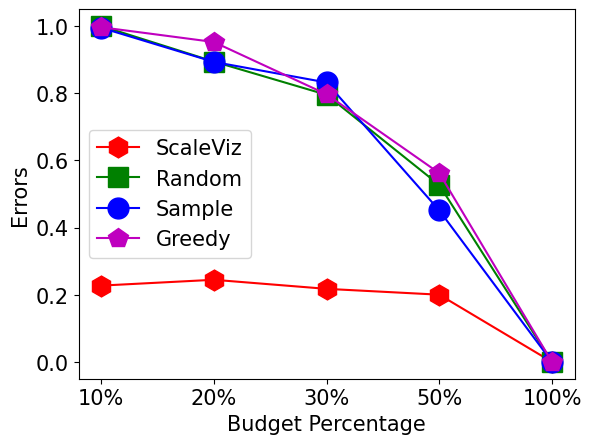}
         \caption{Income dataset}
         \label{fig:three sin x}
     \end{subfigure}

    \begin{subfigure}[b]{0.33\textwidth}
         \centering
         \includegraphics[width=\textwidth]{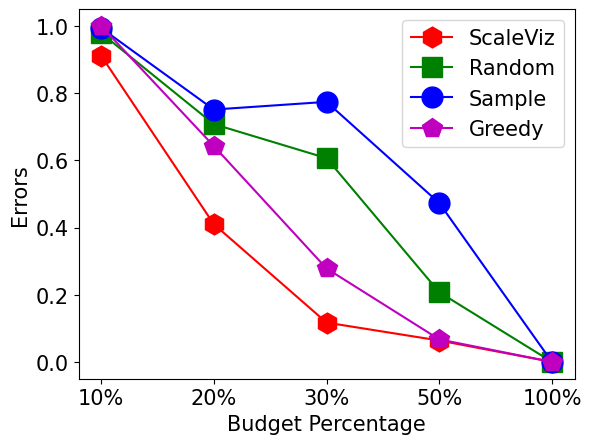}
         \caption{Cars dataset.}
         \label{fig:y equals x}
     \end{subfigure}
     \begin{subfigure}[b]{0.33\textwidth}
         \centering
         \includegraphics[width=\textwidth]{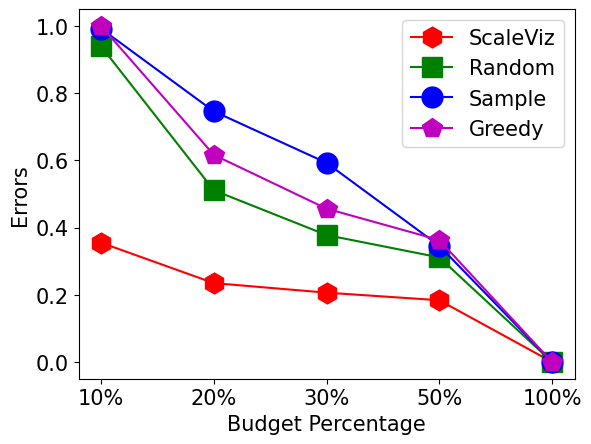}
         \caption{Housing dataset}
         \label{fig:three sin x}
     \end{subfigure}

    \caption{Evaluation of VizML on different datasets}
        \label{fig:three graphs}
\end{figure}

\vspace{-2em}

 
\begin{figure}[H]
     \centering
     \begin{subfigure}[b]{0.33\textwidth}
         \centering
         \includegraphics[width=\textwidth]{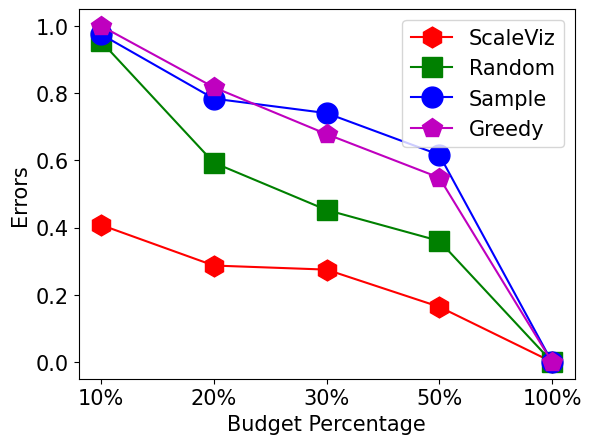}
         \caption{Flights dataset.}
         \label{fig:y equals x}
     \end{subfigure}
     \begin{subfigure}[b]{0.33\textwidth}
         \centering
         \includegraphics[width=\textwidth]{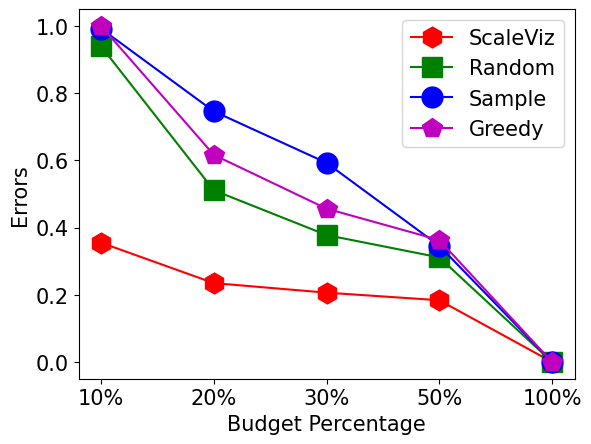}
         \caption{Income dataset}
         \label{fig:three sin x}
     \end{subfigure}

    \begin{subfigure}[b]{0.33\textwidth}
         \centering
         \includegraphics[width=\textwidth]{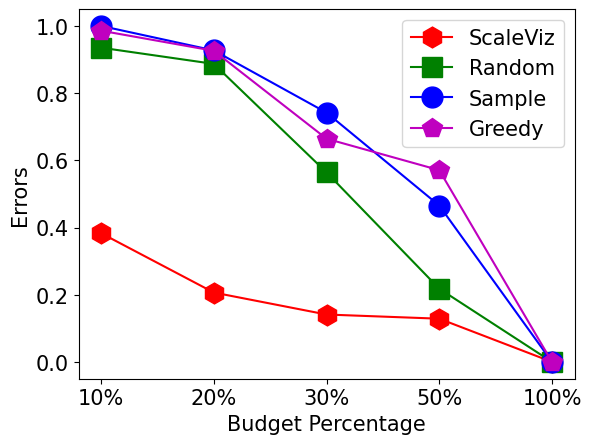}
         \caption{Cars dataset.}
         \label{fig:y equals x}
     \end{subfigure}
     \begin{subfigure}[b]{0.33\textwidth}
         \centering
         \includegraphics[width=\textwidth]{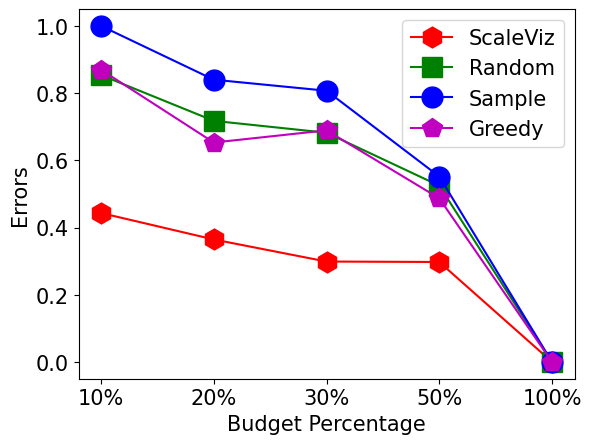}
         \caption{Housing dataset}
         \label{fig:three sin x}
     \end{subfigure}
    
    \caption{Evaluation of MLVR on different datasets}
        \label{fig:four graphs}
\end{figure}

\subsection{Speed-up in Visualization Generation}

We first evaluate the speed-up achieved by \name compared to the baselines approaches when we ensure that the resulting error due to use of less features (for \name, \Random, and \Greedy) or less data (for \Sample) is less than $0.0002, 3.43e^{-05}$ for VizML and MLVR respectively. 
Table~\ref{table:spedup} presents the speedup for four diverse datasets with two target \visrec models. As can be observed that \name helps both the models to choose most effective features, tailored for each datasets, leading to generation of visual recommendtion generation upto 10.3 times faster, which is much higher than the baseline models.





\begin{table}[H]
    \centering
    \begin{tabular}{|l|l|l|l|l|l|l|l|l|}
        \hline
        {} &  \multicolumn{4}{c|}{\textbf{ViZML}} & \multicolumn{4}{c|}{\textbf{MLVR}} \\ \hline
         \textbf{Method} & \textbf{Flights} & \textbf{Income} & \textbf{Housing} & \textbf{Cars} & \textbf{Flights} & \textbf{Income} & \textbf{Housing} & \textbf{Cars} \\ \hline
         \Sample & 1.30 & 1.50 & 1.38 & 1.10 & 1.40 & 2.80 & 1.52 & 2.20 \\ 
        \Greedy & 1.60 & 1.20 & 1.20 & 1.90 & 1.90 & 2.76 & 1.37 & 1.40 \\ 
         \Random & 2.50 & 1.25 & 1.96 & 1.63 & 2.80 & 3.10 & 2.10 & 2.80\\ 
         \name & \textbf{10.3} & \textbf{9.70} & \textbf{10.10} & \textbf{8.10} & \textbf{9.84} & \textbf{9.88} & \textbf{8.60} & \textbf{9.94}\\ \hline
    \end{tabular}
\caption{Speedup in visualization recommendation generation provided by different techniques with limiting errors of $0.0002$ and $3.43e^{-05}$ for VizML and MLVR respectively, compared to results using all the features.}
    \label{table:spedup}
\end{table}

\subsection{Budget vs. Error Trade-off }

We assess the recommendation errors of \name across various budget percentages of the total time on four distinct datasets for two \visrec models, as illustrated in Fig. \ref{fig:three graphs} and Fig. \ref{fig:four graphs}. The errors in Fig, \ref{fig:three graphs} and \ref{fig:four graphs} are normalized to show the difference in errors on a standard scale.  Notably, \name consistently outperforms baselines, showcasing significantly lower errors in visualization recommendations. This effect is particularly prominent at lower budget ranges, highlighting \name's capability to identify the set of most important statistical features that can be computed under a given time-budget constraint while minimizing respective errors for the corresponding base \visrec models. 




\subsection{Need for Dataset-Specific Feature Selection}
We now analyze if there is indeed a need for dataset specific feature-selection. For this, we investigate how much overlap there is in terms of the selected statistical features from different datasets after the runtime feature selection by \name's RL-agent converges to a negligible error with respect to the baseline \visrec models. 
In Table~\ref{tbl:feature_overlap} we show the \textit{intersection over union} (IoU) between the sets of features important features selected by \name for all pairs from the 4 real world datasets. It can be observed that IoU values ranges from 10\% to a maximum of 22\% for VizML. Similarly, for MLVR, the overlap varies from 3\% to a maximum of 14\%. 
This emphasizes the design choice of \name highlighting the fact that feature selection is highly dependent on both the choice of \visrec model ($\mathcal{P}$) and the target dataset ($\mathcal{D}$) and a dataset agnostic pruning of features (even when done in a computation cost-aware manner) would remain suboptimal.

\begin{table}[ht]
  \begin{minipage}{.49\textwidth}
    \centering
    \scalebox{0.9}{
    \begin{tabular}{|c|c|c|c|c|}
    \hline
         & \textbf{Flights} & \textbf{Income} & \textbf{Cars} & \textbf{Housing} \\ \hline
        \textbf{Flights} & 1.00 & 0.22 & 0.10 & 0.12 \\ \hline
        \textbf{Income} & 0.22 & 1.00 & 0.12 & 0.12 \\ \hline
        \textbf{Cars} & 0.10 & 0.12 & 1.00 & 0.19 \\ \hline
        \textbf{Housing} & 0.12 & 0.12 & 0.19 & 1.00 \\ \hline
    \end{tabular}
    }
    \caption{VizML model}
  \end{minipage}%
  \hfill
  \begin{minipage}{.49\textwidth}
    \centering
    \scalebox{0.9}{
    \begin{tabular}{|c|c|c|c|c|}
    \hline
         & \textbf{Flights} & \textbf{Income} & \textbf{Cars} & \textbf{Housing} \\ \hline
        \textbf{Flights} & 1.00 & 0.03 & 0.14 & 0.11 \\ \hline
        \textbf{Income} & 0.03 & 1.00 & 0.03 & 0.05 \\ \hline
        \textbf{Cars} & 0.14 & 0.03 & 1.00 & 0.12 \\ \hline
        \textbf{Housing} & 0.11 & 0.05 & 0.12 & 1.00 \\ \hline
    \end{tabular}
    }
    \caption{MLVR model}
  \end{minipage}
  \caption{Intersection over Union (IoU) of features selected by \name for different datasets. It can be observed that the important statistical features identified for each dataset has very low overlap with other datasets, highlighting the importance of runtime and data-specific feature selection by \name and the fact that a generic feature selection technique would be sub-optimal.}
  \label{tbl:feature_overlap}
\end{table}

\subsection{Scalability with Increasing Data Size}
We now show how \name's benefit keeps on increasing as the size of the dataset (in terms of number of rows) increases. 
We define a saturation budget $\mathcal{B}'$ as the computation time taken by the selected features by \name where the resulting visualization recommendations has insignificant error ( $ \le \epsilon$ ) compared to the base \visrec model. 
For VizML $\epsilon = 0.0002$ and for MLVR $\epsilon = 3.43e^{-05}$.
We us $\mathcal{B}_{MAX}$ to denote the time taken by the base \visrec model to produce the visualizations. 
Table \ref{table:saturate} shows the values of $\mathcal{B}_{MAX}$ and $\mathcal{B}'$ for both VizML and MLVR models for increasing sizes of a dataset (Flights). As can be observed \name saturated at around \textbf{half} the budget for a 1k dataset, saturated at around \textbf{one-fifth} of the budget for 100k, and its efficiency scales even more impressively with larger datasets, reaching about \textbf{one-tenth} of the budget for a dataset size of 1M. This scalability advantage positions \name as an efficient and cost-effective solution to boost \visrec models for large datasets.

\begin{table}[!h]
    \centering
    \begin{tabular}{|l|l|l|l|l|l|l|l|l|}
        \hline
        \visrec & \multicolumn{4}{c|}{\textbf{VizML}} & \multicolumn{4}{c|}{\textbf{MLVR}} \\ \hline
        \textbf{Records} & \textbf{1k} & \textbf{10k} & \textbf{100k} & \textbf{1M} & \textbf{1k} & \textbf{10k} & \textbf{100k} & \textbf{1M} \\ \hline
        \textbf{$\mathcal{B}_{MAX}$} & 160 & 233 & 1311 & 13259 & 592 & 928 & 1950 & 11161 \\
        \textbf{$\mathcal{B'}$} & 79 & 112 & 262 & 1285 & 458 & 583 & 771 & 1134 \\ \hline
    \end{tabular}
    \caption{Analysis of the minimum budget $\mathcal{B'}$ and $\mathcal{B}_{MAX}$ (\textit{milliseconds} for VizML, \textit{seconds} for MLVR) required to achieve specified errors on the flights dataset. VizML to achieve an error $\epsilon = 0.0002$ MLVR to achieve an error $\epsilon = 3.43e^{-05}$ }
    \label{table:saturate}
\end{table}




\section{Conclusion}
In this paper, we identify an important drawback of the state-of-the-art visualization recommendation (\visrec) models that these models sacrificed the scalability in order to make them generalize over unknown datasets. Such models compute a very large number of statistics from the target dataset, which becomes infeasible at larger dataset sizes.  In this paper, we propose \name - a scalable and time-budget-aware framework for visualization recommendations on large datasets. Our approach can be used with existing \visrec models to tailor them for a target dataset, such that visual insights can be generated in a timely manner with insignificant error compared to alternate baseline approaches. 

\label{sec:conclusion}

\bibliographystyle{splncs04}
\bibliography{paper}

\end{document}